\documentclass[letterpaper]{article} % DO NOT CHANGE THIS

\usepackage{aaai2026}              % Camera-ready version
\usepackage{amssymb}
\usepackage{times}  % DO NOT CHANGE THIS
\usepackage{helvet}  % DO NOT CHANGE THIS
\usepackage{courier}  % DO NOT CHANGE THIS
\usepackage[hyphens]{url}  % DO NOT CHANGE THIS
\usepackage{graphicx} % DO NOT CHANGE THIS
\usepackage{amsmath} % Added for \text in math mode
\usepackage{natbib}  % DO NOT CHANGE THIS AND DO NOT ADD ANY OPTIONS TO IT
\usepackage{caption} % DO NOT CHANGE THIS AND DO NOT ADD ANY OPTIONS TO IT
\usepackage[utf8]{inputenc} % allow utf-8 input
\usepackage[T1]{fontenc}    % use 8-bit T1 fonts
\usepackage{url}            % simple URL typesetting
\usepackage{booktabs}       % professional-quality tables
\usepackage{amsfonts}       % blackboard math symbols
\usepackage{nicefrac}       % compact symbols for 1/2, etc.
\usepackage{microtype}      % microtypography
\usepackage{xcolor}         % colors
\usepackage[table]{xcolor}
\usepackage{multirow} 
\usepackage{amsmath}
\usepackage{pifont}
\usepackage{graphicx}
\usepackage{booktabs}
\usepackage{graphicx}
\usepackage{algorithm}
\usepackage{algorithmic}
\usepackage{newfloat}
\usepackage{listings}
\DeclareCaptionStyle{ruled}{labelfont=normalfont,labelsep=colon,strut=off} % DO NOT CHANGE THIS
\floatstyle{ruled}
\newfloat{listing}{tb}{lst}{}
\floatname{listing}{Listing}

\nocopyright 
    \title{MUVLA: Learning to Explore Object Navigation via Map Understanding}
\author{
    Peilong Han\textsuperscript{\rm 1}\thanks{This work was done during the internship at Dexmal}\equalcontrib, 
    \quad Fan Jia\textsuperscript{\rm 2}\equalcontrib, 
    \quad Min Zhang\textsuperscript{\rm 1}, 
    \quad Yutao Qiu\textsuperscript{\rm 3}, 
    \quad Hongyao Tang\textsuperscript{\rm 1}, 
    \quad Yan Zheng\textsuperscript{\rm 1}, 
    \quad Tiancai Wang\textsuperscript{\rm 2}, 
    \quad Jianye Hao\textsuperscript{\rm 1}\thanks{Corresponding author}
}
\affiliations{
    \textsuperscript{\rm 1}Tianjin University 
    \quad \textsuperscript{\rm 2}Dexmal
    \quad \textsuperscript{\rm 3}Beijing Institute of Technology
}

\usepackage{bibentry}
\begin{document}

\maketitle

\begin{abstract}
In this paper, we present MUVLA, a Map Understanding Vision-Language-Action model tailored for object navigation. It leverages semantic map abstractions to unify and structure historical information, encoding spatial context in a compact and consistent form. MUVLA takes the current and history observations, as well as the semantic map, as inputs and predicts the action sequence based on the description of goal object. Furthermore, it amplifies supervision through reward-guided return modeling based on dense short-horizon progress signals, enabling the model to develop a detailed understanding of action value for reward maximization.
MUVLA employs a three-stage training pipeline: learning map-level spatial understanding, imitating behaviors from mixed-quality demonstrations, and reward amplification. This strategy allows MUVLA to unify diverse demonstrations into a robust spatial representation and generate more rational exploration strategies.
Experiments on HM3D and Gibson benchmarks demonstrate that MUVLA achieves great generalization and learns effective exploration behaviors even from low-quality or partially successful trajectories.
\end{abstract}

% Version-specific content
\ifdefined\aaaianonymous
\section{Preparing an Anonymous Submission}

This document details the formatting requirements for anonymous submissions. The requirements are the same as for camera ready papers but with a few notable differences:

\begin{itemize}
    \item Anonymous submissions must not include the author names and affiliations. Write ``Anonymous Submission'' as the ``sole author'' and leave the affiliations empty.
    \item The PDF document's metadata should be cleared with a metadata-cleaning tool before submitting it. This is to prevent leaked information from revealing your identity.
    \item References must be anonymized whenever the reader can infer that they are to the authors' previous work.
    \item AAAI's copyright notice should not be included as a footer in the first page.
    \item Only the PDF version is required at this stage. No source versions will be requested, nor any copyright transfer form.
\end{itemize}

You can remove the copyright notice and ensure that your names aren't shown by including \texttt{submission} option when loading the \texttt{aaai2026} package:

\begin{quote}\begin{scriptsize}\begin{verbatim}
\documentclass[letterpaper]{article}
\usepackage[submission]{aaai2026}
\end{verbatim}\end{scriptsize}\end{quote}

The remainder of this document are the original camera-ready instructions. Any contradiction of the above points ought to be ignored while preparing anonymous submissions.

\section{Camera-Ready Guidelines}
\else
\section{Introduction}
\fi

% 第一段 on 定义 任务
% 第二段 on 的历史 方法演进 传统方法是小模型 现在方法大模型，但是trainningfree，局限性，需要trainingbased
% 第三段 受到vla vln任务的启发，但是有什么问题vla做on的局限性
% 我们的
Object navigation is an important task in the field of embodied navigation. It requires an agent to locate an instance of that object in a previously unseen environment, given a textual description of a target object (e.g., ``bed''). Object navigation task challenge agents to understand unfamiliar environments and autonomously search for specific objects.
Traditionally, object navigation methods are learning-based, typically relying on neural network trained in end-to-end manner~\cite{chang2020semantic, maksymets2021thda}. With the rapid development of large-scale foundation models~\cite{achiam2023gpt, kenton2019bert, touvron2023llama}, some recent advances in object navigation have seen a shift from learning-based methods to training-free approaches. They typically use external memory modules, such as metric maps or topological graphs, and combine with large-scale models to select long-term goals, and then rely on heuristic local policies for low-level action execution~\cite{yu2023l3mvn, wu2024voronav, yokoyama2024vlfm, kuang2024openfmnav}. Consequently, these training-free methods are heavily dependent on external perception and map construction modules, and reveal a clear gap with the potential of foundation models. Such approaches often lack behavioral flexibility, as they do not possess true spatial understanding but instead rely on engineered substitutes.

With the rapid progress of large-scale foundation models, Vision-Language-Action (VLA) models that have been domain-adapted using high-quality robotic data are demonstrating remarkable potential~\cite{black2410pi0, li2024cogact}. Unlike manipulation tasks in robotics, embodied navigation places greater emphasis on an agent’s perceptual and memory capabilities, requiring sophisticated reasoning and efficient exploration. Meanwhile, some recent learning-based VLA approaches mainly focus on Vision-Language Navigation (VLN), leveraging the language generalization ability of foundation models through supervised fine-tuning (SFT)~\cite{zhang2024navid, zhang2024uni, cheng2024navila, zhang2025mapnavnovelmemoryrepresentation} and reinforcement fine-tuning (RFT)~\cite{gao2025octonav, qi2025vln} with ground-truth actions.

\begin{figure}[t]
  \centering
  \includegraphics[width=\columnwidth]{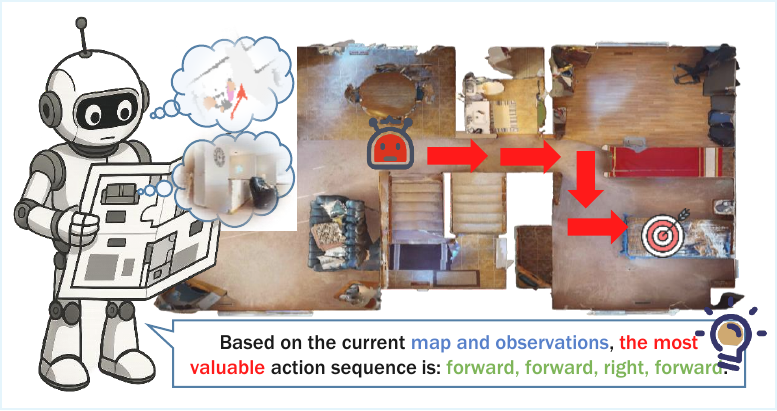}
  \caption{MUVLA focuses on learning efficient exploration strategies by (1) leveraging map-based abstraction to unify diverse and noisy historical trajectories into a robust decision foundation, and (2) learning to evaluate the quality of candidate actions, enabling direct prediction of high-quality actions for efficient navigation.}
  \label{fig1}
\end{figure}
% difference vln引用
Importantly, there is a foundational distinction between VLN, where agents ``follow instructions,'' and object navigation, which requires agents to ``explore independently.'' Crucially, for object navigation, it is challenging to evaluate the optimal trade-off between exploration and efficiency, making it difficult to define what constitutes ground-truth actions—both for use as supervision in behavior cloning and as a reliable evaluation standard for reward modeling in reinforcement fine-tuning. This fundamental difference means that object navigation agents must discover effective exploration strategies beyond simply replicating a fixed language-action mapping.
As a result, several critical questions arise:
% 如何解决xx问题
(1) How to represent historical information in a unified and consistent manner. (2) How to learn an effective and efficient exploration policy from a dataset with mixed-quality demonstrations.
In practice, it is often unclear which trajectories constitute high-quality expert behavior. Relying solely on shortest-path supervision can suppress the agent's exploratory capacity. Conversely, using exploratory trajectories introduces quality variance and increases the burden of both memory and decision-making, as the agent may end up imitating aimless or suboptimal behavior.

In this paper, we present MUVLA, a Map Understanding Vision-Language-Action model. Our approach focuses on training an effective VLA model for object navigation from two aspects: (1) learning a unified historical representation from diverse and noisy past trajectories as a robust basis for decision-making; and (2) extracting high-quality exploration strategies from mixed-quality data (see Fig.~\ref{fig1}). 
To achieve these goals, we adopt a three-stage training pipeline. Firstly, we learn map-level spatial understanding to capture the underlying spatial context. Secondly, we imitate behaviors from demonstrations of varying quality to enrich the model  experience. Thirdly, we apply reward amplification to guide the model towards more effective exploration strategies.

Specifically, we abstract the historical observations into a semantic map, unifying redundant and inconsistent exploration memories into a structured map representation. The semantic map, together with the current and several historical frames, serves as the primary inputs of MUVLA model. Unlike MapNav~\cite{zhang2025mapnavnovelmemoryrepresentation}, we construct a training dataset composed of both rule-generated map descriptions and model-generated chain-of-thought (CoT) reasoning~\cite{wei2022chain}, in order to instill spatial reasoning capabilities during training.
Next, leveraging the map-understanding capacity, MUVLA is trained with paired semantic map and observation frames for behavior cloning. The visual encoder is carefully designed to maximize information utilization while reducing computational overhead.
Finally, we further enhance the VLA model through reward-guided training. Inspired by previous work~\cite{zhuang2024reinformer, zhang2025reinbot}, we introduce a reward head to predict the maximum short-term cumulative return (Return-to-Go~\cite{chen2021decision}) under the current conditions. To ensure the predicted return aligns with the achievable maximum for each goal and state, we employ expectile regression~\cite{sobotka2012geoadditive, aigner1976estimation} as the learning objective. The final loss combines the expectile regression with a reward-weighted behavior cloning loss, encouraging the policy distribution to shift toward higher-quality actions. During inference, the reward head enables MUVLA to predict the maximum attainable return, thereby guiding the agent to execute more robust and effective exploration strategies. In summary, our contributions are three-fold:

\begin{itemize}
\item We propose a unified semantic map abstraction that encodes historical information and leverages reward signals to gain deep insights into data quality, thereby enabling more robust learning and reducing the impact of noisy or redundant trajectories.
\item We introduce a novel three-stage training pipeline that sequentially optimizes map understanding, behavior cloning, and reward-guided policy learning.
\item Extensive experiments on the HM3D and Gibson benchmark show that MUVLA achieves state-of-the-art performance and robust exploration policies among training-based methods, even when trained on heterogeneous-quality demonstration data.
\end{itemize}
% 贡献点

\section{Related Work}
\subsection{Foundation Models for Embodied Navigation}
Recently, an increasing number of works have explored incorporating LLMs and MLLMs into embodied navigation tasks, leveraging the strong generalization abilities of large models for efficient task completion\cite{huang2022visual, zheng2024towards, zhou2024navgpt}. However, the requirements and key challenges differ substantially across tasks. For VLN (Vision-Language Navigation), the primary focus is on aligning language and action, taking advantage of the inherent language generalization of large models. High-quality expert datasets can be collected and expanded as instruction-trajectory pairs for effective supervised alignment and training\cite{gao2025octonav, qi2025vln}.

In contrast, Object Navigation requires an agent to locate an instance of that object in a previously unseen environment. The inherent need for autonomous exploration in such tasks makes it difficult to define what constitutes a “high-quality dataset.” On one hand, prior training-based methods seldom leverage large models for direct action generation during exploration~\cite{chang2020semantic, maksymets2021thda, ye2021auxiliary, mousavian2019visual, yang2018visual}. On the other hand, many training-free approaches merely use pretrained models to assist in waypoint selection~\cite{chang2023goat, yu2023l3mvn, shah2023navigation}, and then rely on the integration of path-planning algorithms~\cite{wu2024voronav}, complementary mapping techniques~\cite{long2024instructnav, yokoyama2024vlfm}, or auxiliary tools~\cite{zhang2024trihelper} to complete the navigation. Some works have also considered unifying multiple embodied tasks within a single framework. However, such approaches require large-scale, high-quality multi-task datasetsand typically rely on video-based models, which impose substantial storage and computational demands during inference\cite{gao2025octonav, zhang2024uni}. In contrast, we targets object navigation by replacing video histories with abstract semantic maps, and further equips the model with the ability to learn efficient exploration strategies directly from mixed-quality data.

\subsection{Foundation Models Integrating with RL}
Recent works have begun to combine reinforcement learning (RL) with foundation models to further improve their accuracy and reasoning abilities\cite{shao2024deepseekmath, lin2025cppo, ramesh2024group}. On one hand, DeepSeek\cite{guo2025deepseek} has shown that even outcome-only rewards can guide LLMs to develop reasoning behaviors without step-level annotations, demonstrating that RL can enhance reasoning skills\cite{mark2024policy, zhai2024fine, guo2025improving}. However, such RL post-training mainly serves to improve data efficiency rather than fundamentally expanding the model’s capabilities\cite{yue2025does}, and is thus less applicable to exploration tasks like object navigation.

On the other hand, inspired by the principles of offline RL, many studies have focused on improving the data efficiency of collected datasets without requiring online data collection\cite{levine2020offline}. Many of these have investigated the use of sequence models as agent policies in decision-making tasks\cite{yamagata2023q, janner2021offline, shafiullah2022behavior}, where the policy model is trained on offline datasets in a supervised manner—conditioned on historical observations and ReturnToGo (RTG)—to predict appropriate actions\cite{chen2021decision, zhuang2024reinformer, zhang2025reinbot}.
We try to explore the integration of RL-based methods into VLA models for object navigation. Specifically, we incorporate the RL principle of return maximization into a general-purpose VLA framework. This enables our approach to make efficient use of mixed-quality demonstration data and more accurately capture the value of individual actions.

\begin{figure*}[t]
  \centering
  \includegraphics[width=1\textwidth]{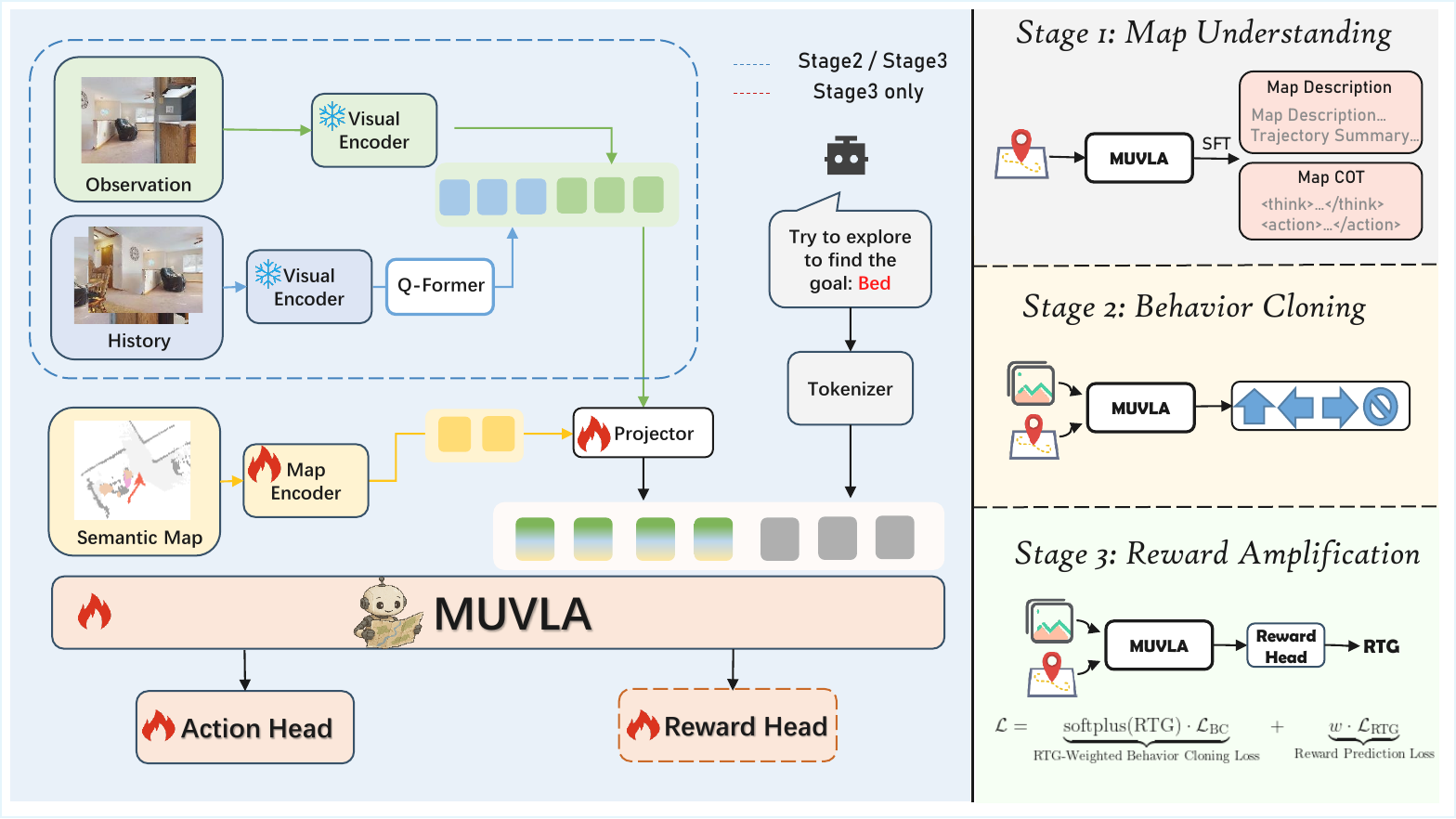}
  \caption{Overview of the MUVLA framework, which integrates semantic maps, observations, and language for efficient object navigation. The three-stage training targets map understanding, behavior cloning, and reward amplification.}
  \label{fig2}
\end{figure*}

\section{Method}
% \subsection{Overview}
In this section, we first present the overall architecture of MUVLA and then provide a detailed description of the training procedure and data preparation pipeline, including the construction of the semantic map and the three-stage training process (as illustrated in Fig.~\ref{fig2}).

\subsection{Model Architecture}
We present the model architecture of MUVLA, which integrates semantic map and visual observation encoding, cross-modal feature fusion, and joint action-reward prediction within a unified vision-language-action framework.
\paragraph{Map and Observation Encoding.}
Given a semantic map image $M_t$ and observation frames  $O$, we encode both modalities using dedicated vision encoders. The semantic map $M_t$ is processed by a map encoder $E_\text{map}$ to extract patch-level features. For the observation modality, the previous three frames and the current frame are encoded by an observation encoder $E_\text{obs}$, with historical features further aggregated by a Q-Former $G_H$. The final map and observation representations are thus:
\begin{equation}
\begin{aligned}
    F_t^{\text{map}} &= E_\text{map}(M_t), \\
    F_t^{\text{obs}} &= \text{Concat}\Big[G_H\big(\{E_\text{obs}(O_{t-i})\}_{i=3}^{1}\big),~ E_\text{obs}(O_t)\Big]
\end{aligned}
\end{equation}

\paragraph{Map-Observation Fusion.}

To enable effective cross-modal reasoning, we fuse the map and observation representations using a cross-attention fusion module $F(\cdot, \cdot)$. Specifically, the observation features serve as the query, while the map features are used as the key and value in the cross-attention mechanism. The fused features are then projected into the multimodal hidden space by a learnable projector $P(\cdot)$ for downstream policy learning. The overall process can be formulated as:
\begin{equation}
\begin{aligned}
    F_t^{\text{fuse}} &= F(F_t^{\text{map}}, F_t^{\text{obs}}), \\
    F_t^{\text{proj}} &= P(F_t^{\text{fuse}})
\end{aligned}
\end{equation}
where $F_t^{\text{fuse}}$ denotes the fused map-observation representation and $F_t^{\text{proj}}$ is the final multimodal embedding used for policy learning.

\paragraph{Action and Reward Prediction.}
Given the projected multimodal representation $F_t^{\text{proj}}$ and the tokenized instruction $\tilde{I}$, the model (LLM) predicts either the action sequence or the reward by applying the corresponding output head:
\begin{align}
    a_{t:t+k-1} &= \mathcal{F}_{\text{action}}\big(\mathrm{LLM}(F_t^{\text{proj}},\, \tilde{I})\big) \\
    \hat{r}_{t} &= \mathcal{F}_{\text{reward}}\big(\mathrm{LLM}(F_t^{\text{proj}},\, \tilde{I})\big)
\end{align}
where $\tilde{I}$ represents the tokenized instruction, and $\mathcal{F}_{\text{action}}$ and $\mathcal{F}_{\text{reward}}$ denote the action and reward heads, respectively.

The action head $\mathcal{F}_{\text{action}}$ is used throughout all three training stages as well as during inference. In contrast, the reward head $\mathcal{F}_{\text{reward}}$ is only introduced in the third training stage to provide reward amplification supervision, and is discarded during inference.

\begin{figure*}[t]
    \vspace{-0.1cm}
  \centering
  \includegraphics[width=1\textwidth]{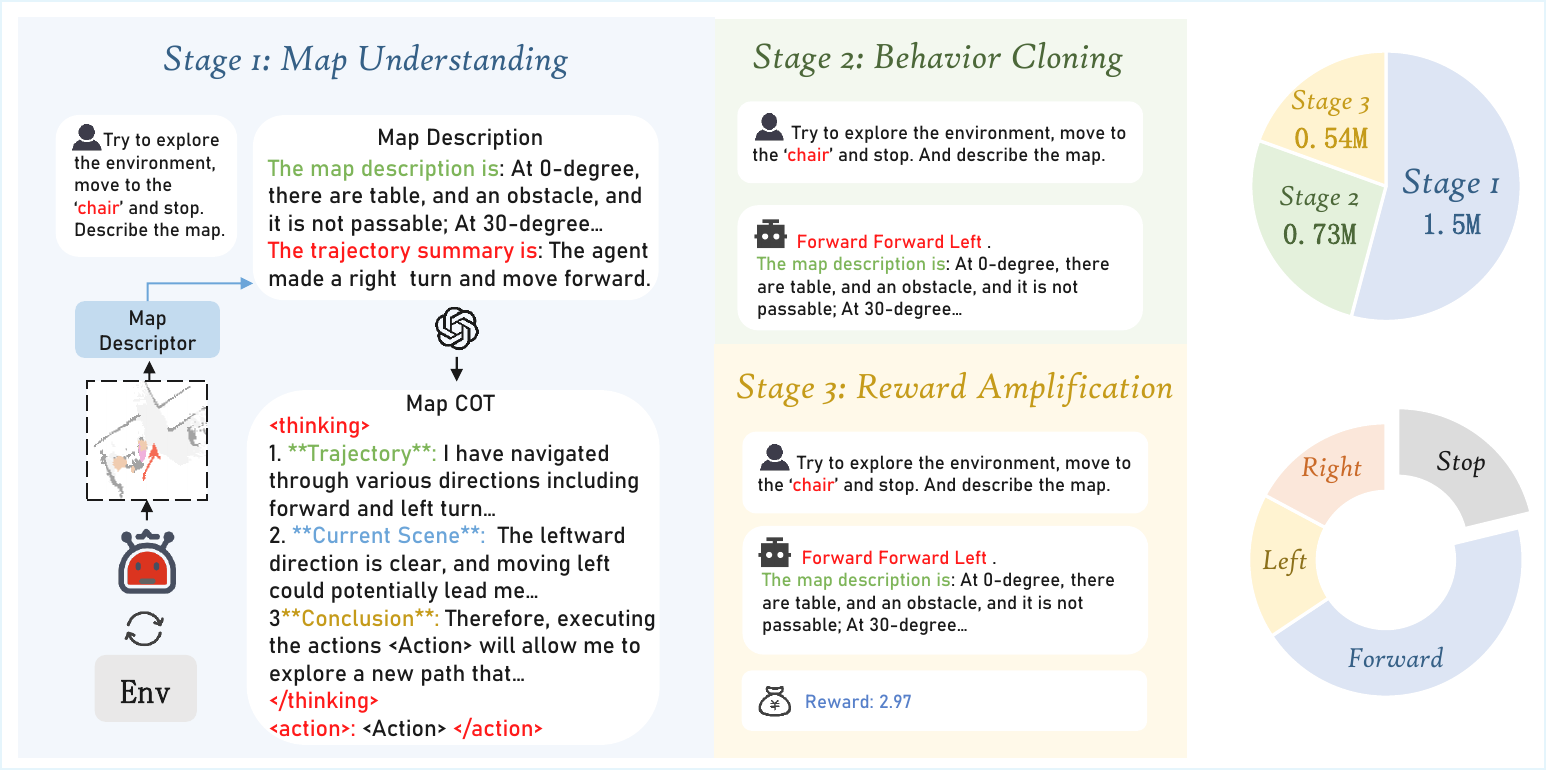}
  \vspace{-0.1cm}
  \caption{Data collection workflow and formats for MUVLA’s three-stage training (left), data volume proportion for each stage (top right), and action distribution in the dataset (bottom right; note that the “stop” action has been augmented).}
  \vspace{-0.2cm}
  \label{fig3}
\end{figure*}

\subsection{Data Collection and Training Recipe}

We employ a structured and comprehensive data collection and training strategy, clearly divided into three sequential stages designed to effectively train the MUVLA model.

\paragraph{Semantic Map Construction.}
To reduce computational and decision-making overhead, and to alleviate the reasoning challenges introduced by redundant exploration, we construct multi-dimensional semantic value maps represented as a \( K \times M \times M \) tensor, where \( M \times M \) denotes the spatial map size and \( K = C + 2 \) is the number of channels. Specifically, \( C \) corresponds to the number of semantic categories, and the first two channels indicate navigable areas and obstacles, respectively.

This \( C + 2 \)-channel semantic map is constructed following the approach in~\cite{chaplot2020object}. At each time step, given RGB-D images and the agent's pose, 3D point clouds are extracted and projected onto the semantic map via height-based filtering. The map is initialized with the agent at the center and is dynamically updated as the agent explores the environment. Point clouds are generated from visual inputs using a geometric pipeline and mapped to a top-down 2D representation. The semantic map contains channels for obstacles and explored regions (derived from depth), as well as semantic channels (from semantic segmentation). Semantic masks are aligned with the corresponding point clouds and projected onto the appropriate map locations. Altogether, these \( C + 2 \) channels provide a comprehensive, persistent record of the environment, capturing navigable areas, obstacles, and semantic categories throughout exploration.

Since an untrained model cannot directly interpret map images, we further simplify the interpretation process and reduce cognitive load by dynamically enlarging and rotating the semantic maps so that the arrow indicating the agent's current position always points upwards as the agent turns. This design mirrors the intuitive experience of human navigation apps, ensuring that the agent's heading is consistently aligned for easier spatial reasoning and decision-making.

\paragraph{Stage 1: Map Understanding.}
To enable the model to acquire spatial understanding abilities prior to downstream policy learning, we first conduct a dedicated pre-training stage focused on semantic map comprehension.
In this stage, we pre-train the model to embed semantic maps into a unified representation space. Specifically, we employ L3MVN~\cite{yu2023l3mvn} as the baseline exploration algorithm within the Habitat simulator on the HM3D dataset, which is also used for subsequent data collection. 

During exploration, at each time step, we record the semantic map image $M_t$, the corresponding textual map information $I_t$, and the agent's action $A_t$. For each step, we construct a map description $D_t$, where $D_t = \{I_t, \{a_0, ..., a_t\}\}$. Here, $I_t$ provides an approximate description of the map in all directions, including objects, obstacles, traversable regions, and the trajectory up to time $t$. 

In addition, we leverage a large language model (LLM) to generate chain-of-thought (CoT) reasoning about map understanding from $D_t$, further enriching the supervision signal for map pre-training. Based on the collected $D_t$, we prompt the LLM to generate a step-wise reasoning process that includes a description of the agent's past trajectory, the current scene, and a high-level conclusion. This structured thinking process is then used to guide the generation of the action sequence $\{a_0, ..., a_t\}$, providing stronger semantic alignment between map descriptions, reasoning, and action labels(see Fig.~\ref{fig3}). 

We collect a total of 1.5M map–language pairs and jointly optimize the map encoder, the projector, and fully fine-tune the LLM, while keeping the observation encoder frozen.

\paragraph{Stage 2: Behavior Cloning.}
To learn an navigation policy grounded in the previously acquired map understanding, we adopt behavior cloning as the second stage of training.
In this stage, each action step is represented as a training sample $(M_t, O, I, A)$. This tuple includes the semantic map $M_t$ at time $t$, multi-modal observations $O$, navigation instructions $I$, and actions $A$. Observations consist of historical frames $O_h \in \mathbb{R}^{3 \times H \times W \times 3}$ from the previous three timesteps and the current observation frame $O_c \in \mathbb{R}^{H \times W \times 3}$. Action labels include the next four steps of navigation, along with the corresponding map description $D_t$, which is incorporated to enhance data diversity and prevent overfitting.

To ensure a balanced distribution of action types, we apply explicit data augmentation, with a particular focus on increasing the frequency of \texttt{stop} actions.
During this stage, the map encoder trained in the first stage is frozen to preserve spatial representations. We also introduce a pre-trained observation encoder, which is likewise frozen during training. All other modules are updated.

With a dataset of 0.73M samples, we define the behavior cloning objective as the conditional action prediction likelihood, optimized via the following cross-entropy loss:
\begin{multline}
    \mathcal{L}_{\text{BC}}(\theta) = 
    -\mathbb{E}_{(M_t, O, I, A) \sim \mathcal{D}_{\text{act}}} \\
    \Bigg[\frac{1}{|A|} \sum_{t=1}^{|A|} \log \pi_\theta \Big(
    A_t \mid\, M_t,\, O,\, I,\, A_{<t}
    \Big) \Bigg]
\end{multline}
where $\pi_\theta$ represents the policy parameterized by weights $\theta$, and $A_{<t}$ denotes previously executed actions.

\paragraph{Stage 3: Reward Amplification.}
To further enhance the model's ability to evaluate action values and prioritize efficient exploration, we introduce explicit reward modeling in the final stage of training. 
Building upon the policy learned in Stage~2, this stage follows the same parameter freezing strategy—keeping both the map encoder and observation encoder frozen, while updating all other modules. Meanwhile, a reward head is newly introduced to support return modeling. This stage is trained using an additional dataset of 0.54M samples specifically collected for reward supervision.

Action rewards $R_t$ are computed based on the reduction in distance to the goal after executing four consecutive actions. For consistent reward scaling across episodes, z-score normalization is applied as follows:
\begin{equation}
R_t = \frac{(d_t - d_{t+4}) - \mu_{\text{ep}}}{\sigma_{\text{ep}}}
\end{equation}
where $d_t$ and $d_{t+4}$ indicate the distances to the goal at steps $t$ and $t+4$, respectively, and $\mu_{\text{ep}}$ and $\sigma_{\text{ep}}$ represent the episode-specific mean and standard deviation.

A short-horizon return-to-go (RTG) strategy is adopted to capture efficient navigation behaviors:
\begin{equation}
\mathrm{RTG}_t = \sum_{k=0}^{K-1} \gamma^k R_{t+k}
\end{equation}

where $K$ denotes the short-horizon window size, $\gamma$ is the discount factor. 
The overall training objective combines reward-weighted action loss and reward prediction loss:
\begin{equation}
    \mathcal{L} = \mathrm{softplus}\big(\mathrm{RTG}_t\big) \cdot \mathcal{L}_\text{BC} + \lambda\, \mathcal{L}_\text{RTG}
\end{equation}

where $\lambda$ is a balancing hyperparameter.

The reward prediction loss $\mathcal{L}_\text{RTG}$ is defined as an expectile regression objective:
\begin{equation}
    \mathcal{L}_\text{RTG} = \mathbb{E}_\tau \left[\, \big| \tau - \mathbb{I}(\Delta g < 0) \big|\, (\Delta g)^2 \,\right]
\end{equation}
where $\Delta g = \hat{\mathrm{RTG}} - \mathrm{RTG}$, $\tau$ is the expectile parameter, and $\mathbb{I}(\cdot)$ denotes the indicator function.

Here, $\mathbb{I}(\cdot)$ is a binary indicator function that outputs $1$ if its argument is true and $0$ otherwise, and $\tau \in (0, 1)$ is a hyperparameter controlling the asymmetry of expectile regression. A larger $\tau$ encourages the model to focus more on high-return samples, thereby making the model more sensitive to trajectories that achieve higher rewards and improving its ability to maximize returns during navigation.

\section{Experiments}

In this section, we evaluate the performance of MUVLA against existing baselines on the HM3D~\cite{ramakrishnan2021habitat}and Gibsion \cite{xia2018gibson} dataset.

\subsection{Experimental Setup}

\paragraph{Dataset.}
We train our model using the 75 training scenes from the HM3D dataset~\cite{ramakrishnan2021habitat}, which provides diverse and realistic indoor environments for embodied navigation. For evaluation, we use the 20 validation scenes from HM3D, and additionally perform zero-shot testing on the Gibson validation split, which consists of 1,000 episodes across 5 previously unseen scenes.

\paragraph{Metrics.}
To comprehensively assess navigation ability, we adopt two standard metrics: Success Rate (SR) and Success weighted by Path Length (SPL).

\vspace{-1mm}
\begin{itemize}
    \setlength{\itemsep}{0pt}
    \setlength{\parsep}{0pt}
    \setlength{\parskip}{0pt}
    \item \textbf{Success Rate (SR)} quantifies the percentage of episodes in which the agent successfully reaches the target. It is defined as
    \[
        \text{SR} = \frac{1}{N} \sum_{i=1}^{N} S_i,
    \]
    where \( S_i = 1 \) if the episode is successful and \( 0 \) otherwise, and \( N \) is the total number of episodes.
    \item \textbf{Success weighted by Path Length (SPL)} evaluates both success and navigation efficiency, rewarding shorter and more direct trajectories. It is computed as
    \[
        \text{SPL} = \frac{1}{N} \sum_{i=1}^{N} \frac{S_i \cdot l_i}{\max(p_i, l_i)},
    \]
    where \( l_i \) is the shortest path length, \( p_i \) is the agent's actual path length, and \( S_i \) indicates episode success.
\end{itemize}

Higher values of SR and SPL indicate better navigation performance.

\begin{table}[t]
    \centering
    \renewcommand{\arraystretch}{1.2}
    \fontsize{15}{15}\selectfont
    \resizebox{0.95\linewidth}{!}{
    \begin{tabular}{lcccc}
        \toprule
        \multirow{2}{*}{\textbf{Method}} & \multicolumn{2}{c}{\textbf{HM3D}} & \multicolumn{2}{c}{\textbf{Gibson}} \\
        \cmidrule(lr){2-3} \cmidrule(lr){4-5}
        & \textbf{Success$\uparrow$} & \textbf{SPL$\uparrow$} & \textbf{Success$\uparrow$} & \textbf{SPL$\uparrow$} \\
        \midrule
        Random              & 0.00 & 0.00  & 0.03  & 0.03 \\
        SemExp~\cite{chaplot2020object}  & 37.9 & 18.8  & 65.2  & 33.6 \\
        ZSON~\cite{majumdar2022zson}    & 25.5 & 12.6  & --    & --   \\
        Pixel-Nav~\cite{cai2023bridging} & 37.9 & 20.5  & --    & --   \\
        ESC~\cite{zhou2023esc}           & 39.2 & 22.3  & --    & --   \\
        COW~\cite{gadre2023cows}         & 32.0 & 18.1  & --    & --   \\
        FBE~\cite{gervet2023navigating}  & 23.7 & 12.3  & 41.7  & 21.4 \\
        MapNav~\cite{zhang2025mapnavnovelmemoryrepresentation} & 34.6 & \textbf{25.6} & -- & -- \\
        \textbf{MUVLA (Ours)*}            & \textbf{46.7} & 21.0 & \textbf{71.0} & \textbf{41.1} \\
        \bottomrule
    \end{tabular}
    }
    \caption{Comparison with other methods on HM3D and Gibson. The best results are highlighted in \textbf{bold}. \textbf{*} denotes our method evaluated in a zero-shot manner on Gibson, i.e., trained only on HM3D and directly tested on Gibson without any further finetuning. }
    \label{tab1}
\end{table}
% 分成两个表

\subsection{Experimental Results}

As shown in Table~\ref{tab1}, MUVLA achieves the best overall performance among training-based methods on the HM3D benchmark, reaching a Success Rate of 46.7 and an SPL of 21.0. Notably, compared to MapNav~\cite{zhang2025mapnavnovelmemoryrepresentation}, which also leverages map-based memory representations, MUVLA achieves an absolute improvement of $+12.1\%$ in Success Rate. Furthermore, MUVLA achieves strong zero-shot generalization on the Gibson benchmark, obtaining a Success Rate of 71.0 and an SPL of 41.1 without any additional finetuning. This demonstrates that our approach not only excels on the training domain but also transfers effectively to novel, unseen environments. Our approach prioritizes robust and generalizable success across diverse and noisy trajectories, as reflected in the higher Success Rate.

Compared to other baselines such as SemExp~\cite{chaplot2020object}, Pixel-Nav~\cite{cai2023bridging}, and ESC~\cite{zhou2023esc}, MUVLA consistently surpasses them by a large margin on the Success Rate metric, demonstrating the effectiveness of our map-centric abstraction and reward-aware training pipeline. These results highlight MUVLA’s superior capability in learning efficient exploration strategies from mixed-quality data and unifying diverse past experiences for robust navigation.

\begin{table}[t]
    \centering
    \renewcommand{\arraystretch}{1.2}
    \fontsize{15}{15}\selectfont
    \resizebox{0.8\linewidth}{!}{
    \begin{tabular}{ccc|cc}
        \toprule
        \multicolumn{3}{c|}{\textbf{Training Stage}} & \multicolumn{2}{c}{\textbf{Metrics}} \\
        \cmidrule(lr){1-3} \cmidrule(lr){4-5}
        \textbf{Stage 1} & \textbf{Stage 2} & \textbf{Stage 3} & \textbf{Success$\uparrow$} & \textbf{SPL$\uparrow$} \\
        \midrule
         & \checkmark &  & 38.1 & 18.9 \\
         &  & \checkmark & 32.3 & 13.5 \\
         & \checkmark & \checkmark & 40.7 & 18.6 \\
        \checkmark & \checkmark &  & 42.8 & 18.4 \\
        \midrule
        \checkmark & \checkmark & \checkmark & 46.7 & 21.0 \\
        \bottomrule
    \end{tabular}
    }
    \caption{Ablation results of MUVLA on the HM3D validation set. A checkmark indicates that the corresponding training stage is used.}
    \label{tab2}
\end{table}

% 换成打勾

\subsection{Ablation Study}

\paragraph{Impact of Training Pipeline.}
The ablation study results in Table~\ref{tab2} demonstrate the importance of each training stage in MUVLA. Incorporating the dedicated map understanding stage (Stage 1) enables the model to better align abstract semantic maps with language, thereby enhancing its spatial reasoning and interpretation of historical exploration trajectories; omitting this stage (\textit{w/o Stage 1}) still allows the model to use the map as an external modality and achieve reasonable performance (40.7\% Success, 18.6 SPL), but the overall navigation effectiveness is noticeably reduced compared to the full model (46.7\% Success, 21.0 SPL). The behavior cloning stage (Stage 2) is essential for basic navigation competence, but relying solely on imitation learning (\textit{only Stage 2}) further limits the model’s ability to prioritize actions that lead to more efficient exploration (only 38.1\% Success, 18.9 SPL), especially when dealing with mixed-quality demonstration data. The reward amplification stage (Stage 3) further improves the agent’s performance by explicitly incorporating return-based supervision, encouraging the model to focus on actions that maximize cumulative rewards; however, using only Stage 3 (\textit{only Stage 3}) leads to insufficient supervision and inferior results (32.3\% Success, 13.5 SPL) compared to the full pipeline. 

The best performance is achieved when all three stages are combined, highlighting their complementary strengths: map-language alignment brings robust spatial priors, imitation learning captures successful action patterns, and reward-based training promotes value-driven exploration. Together, these components enable MUVLA to learn effective object navigation policies from large-scale, mixed-quality datasets.

\begin{table}[t]
    \centering
    \renewcommand{\arraystretch}{1.2}
    \fontsize{15}{15}\selectfont
    \resizebox{0.98\linewidth}{!}{
    \begin{tabular}{c|c|c|c|cc}
        \toprule
        \multirow{2}{*}{\textbf{Map Input}} & 
        \multicolumn{2}{c|}{\textbf{Stage 1 Data}} & 
        \multirow{2}{*}{\textbf{Stage 3}} & 
        \multicolumn{2}{c}{\textbf{Metrics}} \\
        \cmidrule(lr){2-3} \cmidrule(lr){5-6}
        & \textbf{CoT} & \textbf{Desc.} & & \textbf{Success$\uparrow$} & \textbf{SPL$\uparrow$} \\
        \midrule
                \ & \ & \ &  & 36.5 & 16.5 \\
         & \ & \ & \checkmark & 39.9 & 19.6 \\
        \checkmark &  & \checkmark &  & 43.8 & 19.6 \\
        \checkmark &  & \checkmark & \checkmark & 45.1 & 20.2 \\
        \checkmark & \checkmark &  &  & 23.2 & 10.7 \\
        \checkmark & \checkmark &  & \checkmark & 39.6 & 18.6 \\  
        \checkmark & \checkmark & \checkmark & \checkmark & 46.7 & 21.0 \\
        \bottomrule
    \end{tabular}
    }
    \caption{Ablation study results on the contribution of map input, Stage 1 data (Map CoT and Description). A checkmark indicates the corresponding component is used.}
    \label{tab3}
\end{table}

\paragraph{Impact of Map Understanding Training.}
The ablation results in Table~\ref{tab3} quantitatively demonstrate the critical role of map-based inputs and rich supervision in MUVLA’s navigation performance. Removing the map modality as input causes the Success rate to drop from 46.7\% to 39.9\% and SPL from 21.0 to 19.6, underlining the necessity of semantic maps as a robust context for decision-making.

When most map description data is removed in the first stage (\textit{Stage1 w/o Map Description}), performance drops sharply to 39.6\% Success and 18.6 SPL, a relative decrease of about 15\% in Success compared to the full model, highlighting the importance of abundant map-language pairs. By contrast, removing only a small number of chain-of-thought (COT) annotations (\textit{Stage1 w/o Map COT}) results in only a minor reduction, from 46.7\% to 45.1\% Success and 21.0 to 20.2 SPL, suggesting large-scale descriptive supervision is more influential than a small amount of COT data.
However, when training with only COT data (\textit{Stage1 w/o Map Description, w/o Stage 3}), performance degrades dramatically to 23.2\% Success and 10.7 SPL—representing a drop of more than 40\% compared to the full model. This suggests that small-scale COT supervision alone may cause the model to forget or underutilize its general spatial reasoning abilities.

Most notably, the reward amplification in Stage 3 leads to significant performance boosts, especially in cases with limited COT data, where the addition of Stage 3 improves Success by 16.4 percentage points (from 23.2\% to 39.6\%) and SPL by 7.9 (from 10.7 to 18.6)—an increase of over 70\%. This confirms that reward-based training is essential for efficient exploration and robust navigation, particularly when high-quality supervision is scarce.

\paragraph{Impact of Reward Amplification Training.}
The ablation results in Table~\ref{tab4} quantitatively demonstrate the effect of different reward heads and weighting strategies. Incorporating the reward head with expectile regression improves Success from 44.2\% (without reward head) to 46.7\% (main model), and SPL from 19.6 to 21.0. Removing reward weighting and using standard cross-entropy (Stage3 w/o Reward-Weight) yields a lower Success of 43.2\% and SPL of 18.7, confirming that adaptive weighting based on RTG brings measurable gains.

Replacing RTG with step-wise instant reward (Stage3 with Instant-Reward) achieves 45.6\% Success and 20.2 SPL, which is close but still lower than the main RTG-based setting, suggesting that cumulative rewards better reflect longer-term value. In contrast, exponential reward-weighting leads to complete failure (0.0\% Success, 0.0 SPL), likely due to unstable gradient scaling.
The use of turning-based reward further lowers performance (40.3\% Success, 19.8 SPL), showing that encouraging or penalizing specific action types (e.g., turning) may restrict flexible navigation.

Finally, replacing the third stage with GRPO (31.0\% Success, 11.0 SPL) demonstrates the importance of structured training and reward design. In this setup, the model receives a reward of 0.25 for each of the four predicted actions that matches the ground-truth action in the dataset, resulting in a maximum reward of 1.0 per step. However, this reward is assigned based solely on offline alignment, without considering step-level supervision or online interaction. As a result, the policy tends to overfit to noise or suboptimal behavior in the dataset, leading to a significant performance drop. These findings highlight the necessity of both progressive training and carefully designed reward signals when learning from imperfect offline demonstrations.

\begin{table}[t]
    \centering
    \renewcommand{\arraystretch}{1.2}
    \fontsize{15}{15}\selectfont
    \resizebox{0.95\linewidth}{!}{
    \begin{tabular}{c|c|c|cc}
        \toprule
        \multicolumn{3}{c|}{\textbf{Stage 3 Condition}} & \multicolumn{2}{c}{\textbf{Metrics}} \\
        \cmidrule(lr){1-3} \cmidrule(lr){4-5}
        \textbf{Reward Head} & \textbf{Reward Weight} & \textbf{Reward Type} & \textbf{Success$\uparrow$} & \textbf{SPL$\uparrow$} \\
        \midrule
         & Softplus & RTG & 44.2 & 19.6 \\
        \checkmark & None & RTG & 43.2 & 18.7 \\
        \checkmark & Exp & RTG & 0.0 & 0.0 \\
        \checkmark & Softplus & Instant Reward & 45.6 & 20.2 \\
        \checkmark & Softplus & Turning Reward & 40.3 & 19.8 \\
        \midrule
        \multicolumn{3}{c|}{GRPO} & 31.0 & 11.0 \\
        \midrule
        \checkmark & Softplus & RTG & 46.7 & 21.0 \\
        \bottomrule
    \end{tabular}
    }
    \caption{Ablation study results on the contribution of reward amplification components in Stage 3. A checkmark indicates the corresponding component is enabled.}
    \label{tab4}
\end{table}

\section{Conclusion}
% 修改表格
In this work, we propose MUVLA, a unified Vision-Language-Action framework designed for object navigation. By abstracting historical experiences into multi-dimensional semantic memory maps and leveraging map-language alignment, our method enables spatial reasoning over diverse and noisy past trajectories. We introduce a three-stage training pipeline, incorporating map understanding, behavior cloning, and reward-weighted policy optimization, which collectively allow the model to learn effective exploration strategies from mixed-quality offline demonstrations. Extensive experiments on the HM3D and Gibson benchmark demonstrate that MUVLA significantly outperforms prior learning-based approaches. Ablation studies further validate the importance of each component. Our results highlight the potential of integrating spatial abstraction, language understanding, and value-driven learning for generalizable embodied navigation.

\bibliography{aaai2026}

\begin{thebibliography}{53}
\providecommand{\natexlab}[1]{#1}

\bibitem[{Achiam et~al.(2023)Achiam, Adler, Agarwal, Ahmad, Akkaya, Aleman, Almeida, Altenschmidt, Altman, Anadkat et~al.}]{achiam2023gpt}
Achiam, J.; Adler, S.; Agarwal, S.; Ahmad, L.; Akkaya, I.; Aleman, F.~L.; Almeida, D.; Altenschmidt, J.; Altman, S.; Anadkat, S.; et~al. 2023.
\newblock Gpt-4 technical report.
\newblock \emph{arXiv preprint arXiv:2303.08774}.

\bibitem[{Aigner, Amemiya, and Poirier(1976)}]{aigner1976estimation}
Aigner, D.~J.; Amemiya, T.; and Poirier, D.~J. 1976.
\newblock On the estimation of production frontiers: maximum likelihood estimation of the parameters of a discontinuous density function.
\newblock \emph{International economic review}, 377--396.

\bibitem[{Black et~al.()Black, Brown, Driess, Esmail, Equi, Finn, Fusai, Groom, Hausman, Ichter et~al.}]{black2410pi0}
Black, K.; Brown, N.; Driess, D.; Esmail, A.; Equi, M.; Finn, C.; Fusai, N.; Groom, L.; Hausman, K.; Ichter, B.; et~al. ????
\newblock $\pi$0: A vision-language-action flow model for general robot control. CoRR, abs/2410.24164, 2024. doi: 10.48550.
\newblock \emph{arXiv preprint ARXIV.2410.24164}.

\bibitem[{Cai et~al.(2023)Cai, Huang, Cheng, Long, Gao, Sun, and Dong}]{cai2023bridging}
Cai, W.; Huang, S.; Cheng, G.; Long, Y.; Gao, P.; Sun, C.; and Dong, H. 2023.
\newblock Bridging zero-shot object navigation and foundation models through pixel-guided navigation skill.
\newblock \emph{arXiv preprint arXiv:2309.10309}.

\bibitem[{Chang et~al.(2023)Chang, Gervet, Khanna, Yenamandra, Shah, Min, Shah, Paxton, Gupta, Batra et~al.}]{chang2023goat}
Chang, M.; Gervet, T.; Khanna, M.; Yenamandra, S.; Shah, D.; Min, S.~Y.; Shah, K.; Paxton, C.; Gupta, S.; Batra, D.; et~al. 2023.
\newblock Goat: Go to any thing.
\newblock \emph{arXiv preprint arXiv:2311.06430}.

\bibitem[{Chang, Gupta, and Gupta(2020)}]{chang2020semantic}
Chang, M.; Gupta, A.; and Gupta, S. 2020.
\newblock Semantic visual navigation by watching youtube videos.
\newblock \emph{Advances in Neural Information Processing Systems}, 33: 4283--4294.

\bibitem[{Chaplot et~al.(2020)Chaplot, Gandhi, Gupta, and Salakhutdinov}]{chaplot2020object}
Chaplot, D.~S.; Gandhi, D.~P.; Gupta, A.; and Salakhutdinov, R.~R. 2020.
\newblock Object goal navigation using goal-oriented semantic exploration.
\newblock \emph{Advances in Neural Information Processing Systems}, 33: 4247--4258.

\bibitem[{Chen et~al.(2021)Chen, Lu, Rajeswaran, Lee, Grover, Laskin, Abbeel, Srinivas, and Mordatch}]{chen2021decision}
Chen, L.; Lu, K.; Rajeswaran, A.; Lee, K.; Grover, A.; Laskin, M.; Abbeel, P.; Srinivas, A.; and Mordatch, I. 2021.
\newblock Decision transformer: Reinforcement learning via sequence modeling.
\newblock \emph{Advances in neural information processing systems}, 34: 15084--15097.

\bibitem[{Cheng et~al.(2024)Cheng, Ji, Yang, Gongye, Zou, Kautz, B{\i}y{\i}k, Yin, Liu, and Wang}]{cheng2024navila}
Cheng, A.-C.; Ji, Y.; Yang, Z.; Gongye, Z.; Zou, X.; Kautz, J.; B{\i}y{\i}k, E.; Yin, H.; Liu, S.; and Wang, X. 2024.
\newblock Navila: Legged robot vision-language-action model for navigation.
\newblock \emph{arXiv preprint arXiv:2412.04453}.

\bibitem[{Gadre et~al.(2023)Gadre, Wortsman, Ilharco, Schmidt, and Song}]{gadre2023cows}
Gadre, S.~Y.; Wortsman, M.; Ilharco, G.; Schmidt, L.; and Song, S. 2023.
\newblock Cows on pasture: Baselines and benchmarks for language-driven zero-shot object navigation.
\newblock In \emph{Proceedings of the IEEE/CVF Conference on Computer Vision and Pattern Recognition}, 23171--23181.

\bibitem[{Gao et~al.(2025)Gao, Jin, Peng, Zhang, Deng, Li, Wang, and Liu}]{gao2025octonav}
Gao, C.; Jin, L.; Peng, X.; Zhang, J.; Deng, Y.; Li, A.; Wang, H.; and Liu, S. 2025.
\newblock OctoNav: Towards Generalist Embodied Navigation.
\newblock \emph{arXiv preprint arXiv:2506.09839}.

\bibitem[{Gervet et~al.(2023)Gervet, Chintala, Batra, Malik, and Chaplot}]{gervet2023navigating}
Gervet, T.; Chintala, S.; Batra, D.; Malik, J.; and Chaplot, D.~S. 2023.
\newblock Navigating to objects in the real world.
\newblock \emph{Science Robotics}, 8(79): eadf6991.

\bibitem[{Guo et~al.(2025{\natexlab{a}})Guo, Yang, Zhang, Song, Zhang, Xu, Zhu, Ma, Wang, Bi et~al.}]{guo2025deepseek}
Guo, D.; Yang, D.; Zhang, H.; Song, J.; Zhang, R.; Xu, R.; Zhu, Q.; Ma, S.; Wang, P.; Bi, X.; et~al. 2025{\natexlab{a}}.
\newblock Deepseek-r1: Incentivizing reasoning capability in llms via reinforcement learning.
\newblock \emph{arXiv preprint arXiv:2501.12948}.

\bibitem[{Guo et~al.(2025{\natexlab{b}})Guo, Zhang, Chen, Ji, Wang, Hu, and Chen}]{guo2025improving}
Guo, Y.; Zhang, J.; Chen, X.; Ji, X.; Wang, Y.-J.; Hu, Y.; and Chen, J. 2025{\natexlab{b}}.
\newblock Improving vision-language-action model with online reinforcement learning.
\newblock \emph{arXiv preprint arXiv:2501.16664}.

\bibitem[{Huang et~al.(2022)Huang, Mees, Zeng, and Burgard}]{huang2022visual}
Huang, C.; Mees, O.; Zeng, A.; and Burgard, W. 2022.
\newblock Visual language maps for robot navigation.
\newblock \emph{arXiv preprint arXiv:2210.05714}.

\bibitem[{Janner, Li, and Levine(2021)}]{janner2021offline}
Janner, M.; Li, Q.; and Levine, S. 2021.
\newblock Offline reinforcement learning as one big sequence modeling problem.
\newblock \emph{Advances in neural information processing systems}, 34: 1273--1286.

\bibitem[{Kenton and Toutanova(2019)}]{kenton2019bert}
Kenton, J. D. M.-W.~C.; and Toutanova, L.~K. 2019.
\newblock Bert: Pre-training of deep bidirectional transformers for language understanding.
\newblock In \emph{Proceedings of naacL-HLT}, volume~1, 2.

\bibitem[{Kuang, Lin, and Jiang(2024)}]{kuang2024openfmnav}
Kuang, Y.; Lin, H.; and Jiang, M. 2024.
\newblock OpenFMNav: Towards Open-Set Zero-Shot Object Navigation via Vision-Language Foundation Models.
\newblock \emph{arXiv preprint arXiv:2402.10670}.

\bibitem[{Levine et~al.(2020)Levine, Kumar, Tucker, and Fu}]{levine2020offline}
Levine, S.; Kumar, A.; Tucker, G.; and Fu, J. 2020.
\newblock Offline reinforcement learning: Tutorial, review, and perspectives on open problems.
\newblock \emph{arXiv preprint arXiv:2005.01643}.

\bibitem[{Li et~al.(2024)Li, Liang, Wang, Luo, Chen, Liao, Wei, Deng, Xu, Zhang et~al.}]{li2024cogact}
Li, Q.; Liang, Y.; Wang, Z.; Luo, L.; Chen, X.; Liao, M.; Wei, F.; Deng, Y.; Xu, S.; Zhang, Y.; et~al. 2024.
\newblock Cogact: A foundational vision-language-action model for synergizing cognition and action in robotic manipulation.
\newblock \emph{arXiv preprint arXiv:2411.19650}.

\bibitem[{Lin et~al.(2025)Lin, Lin, Xie, and Ji}]{lin2025cppo}
Lin, Z.; Lin, M.; Xie, Y.; and Ji, R. 2025.
\newblock Cppo: Accelerating the training of group relative policy optimization-based reasoning models.
\newblock \emph{arXiv preprint arXiv:2503.22342}.

\bibitem[{Long et~al.(2024)Long, Cai, Wang, Zhan, and Dong}]{long2024instructnav}
Long, Y.; Cai, W.; Wang, H.; Zhan, G.; and Dong, H. 2024.
\newblock InstructNav: Zero-shot System for Generic Instruction Navigation in Unexplored Environment.
\newblock \emph{arXiv preprint arXiv:2406.04882}.

\bibitem[{Majumdar et~al.(2022)Majumdar, Aggarwal, Devnani, Hoffman, and Batra}]{majumdar2022zson}
Majumdar, A.; Aggarwal, G.; Devnani, B.; Hoffman, J.; and Batra, D. 2022.
\newblock Zson: Zero-shot object-goal navigation using multimodal goal embeddings.
\newblock \emph{Advances in Neural Information Processing Systems}, 35: 32340--32352.

\bibitem[{Maksymets et~al.(2021)Maksymets, Cartillier, Gokaslan, Wijmans, Galuba, Lee, and Batra}]{maksymets2021thda}
Maksymets, O.; Cartillier, V.; Gokaslan, A.; Wijmans, E.; Galuba, W.; Lee, S.; and Batra, D. 2021.
\newblock Thda: Treasure hunt data augmentation for semantic navigation.
\newblock In \emph{Proceedings of the IEEE/CVF International Conference on Computer Vision}, 15374--15383.

\bibitem[{Mark et~al.(2024)Mark, Gao, Sampaio, Srirama, Sharma, Finn, and Kumar}]{mark2024policy}
Mark, M.~S.; Gao, T.; Sampaio, G.~G.; Srirama, M.~K.; Sharma, A.; Finn, C.; and Kumar, A. 2024.
\newblock Policy agnostic rl: Offline rl and online rl fine-tuning of any class and backbone.
\newblock \emph{arXiv preprint arXiv:2412.06685}.

\bibitem[{Mousavian et~al.(2019)Mousavian, Toshev, Fi{\v{s}}er, Ko{\v{s}}eck{\'a}, Wahid, and Davidson}]{mousavian2019visual}
Mousavian, A.; Toshev, A.; Fi{\v{s}}er, M.; Ko{\v{s}}eck{\'a}, J.; Wahid, A.; and Davidson, J. 2019.
\newblock Visual representations for semantic target driven navigation.
\newblock In \emph{2019 International Conference on Robotics and Automation (ICRA)}, 8846--8852. IEEE.

\bibitem[{Qi et~al.(2025)Qi, Zhang, Yu, Wang, and Zhao}]{qi2025vln}
Qi, Z.; Zhang, Z.; Yu, Y.; Wang, J.; and Zhao, H. 2025.
\newblock VLN-R1: Vision-Language Navigation via Reinforcement Fine-Tuning.
\newblock \emph{arXiv preprint arXiv:2506.17221}.

\bibitem[{Ramakrishnan et~al.(2021)Ramakrishnan, Gokaslan, Wijmans, Maksymets, Clegg, Turner, Undersander, Galuba, Westbury, Chang et~al.}]{ramakrishnan2021habitat}
Ramakrishnan, S.~K.; Gokaslan, A.; Wijmans, E.; Maksymets, O.; Clegg, A.; Turner, J.; Undersander, E.; Galuba, W.; Westbury, A.; Chang, A.~X.; et~al. 2021.
\newblock Habitat-matterport 3d dataset (hm3d): 1000 large-scale 3d environments for embodied ai.
\newblock \emph{arXiv preprint arXiv:2109.08238}.

\bibitem[{Ramesh et~al.(2024)Ramesh, Hu, Chaimalas, Mehta, Sessa, Bou~Ammar, and Bogunovic}]{ramesh2024group}
Ramesh, S.~S.; Hu, Y.; Chaimalas, I.; Mehta, V.; Sessa, P.~G.; Bou~Ammar, H.; and Bogunovic, I. 2024.
\newblock Group robust preference optimization in reward-free rlhf.
\newblock \emph{Advances in Neural Information Processing Systems}, 37: 37100--37137.

\bibitem[{Shafiullah et~al.(2022)Shafiullah, Cui, Altanzaya, and Pinto}]{shafiullah2022behavior}
Shafiullah, N.~M.; Cui, Z.; Altanzaya, A.~A.; and Pinto, L. 2022.
\newblock Behavior transformers: Cloning $ k $ modes with one stone.
\newblock \emph{Advances in neural information processing systems}, 35: 22955--22968.

\bibitem[{Shah et~al.(2023)Shah, Equi, Osi{\'n}ski, Xia, Ichter, and Levine}]{shah2023navigation}
Shah, D.; Equi, M.~R.; Osi{\'n}ski, B.; Xia, F.; Ichter, B.; and Levine, S. 2023.
\newblock Navigation with large language models: Semantic guesswork as a heuristic for planning.
\newblock In \emph{Conference on Robot Learning}, 2683--2699. PMLR.

\bibitem[{Shao et~al.(2024)Shao, Wang, Zhu, Xu, Song, Bi, Zhang, Zhang, Li, Wu et~al.}]{shao2024deepseekmath}
Shao, Z.; Wang, P.; Zhu, Q.; Xu, R.; Song, J.; Bi, X.; Zhang, H.; Zhang, M.; Li, Y.; Wu, Y.; et~al. 2024.
\newblock Deepseekmath: Pushing the limits of mathematical reasoning in open language models, 2024.
\newblock \emph{URL https://arxiv. org/abs/2402.03300}, 2(3): 5.

\bibitem[{Sobotka and Kneib(2012)}]{sobotka2012geoadditive}
Sobotka, F.; and Kneib, T. 2012.
\newblock Geoadditive expectile regression.
\newblock \emph{Computational Statistics \& Data Analysis}, 56(4): 755--767.

\bibitem[{Touvron et~al.(2023)Touvron, Lavril, Izacard, Martinet, Lachaux, Lacroix, Rozi{\`e}re, Goyal, Hambro, Azhar et~al.}]{touvron2023llama}
Touvron, H.; Lavril, T.; Izacard, G.; Martinet, X.; Lachaux, M.-A.; Lacroix, T.; Rozi{\`e}re, B.; Goyal, N.; Hambro, E.; Azhar, F.; et~al. 2023.
\newblock Llama: Open and efficient foundation language models.
\newblock \emph{arXiv preprint arXiv:2302.13971}.

\bibitem[{Wei et~al.(2022)Wei, Wang, Schuurmans, Bosma, Xia, Chi, Le, Zhou et~al.}]{wei2022chain}
Wei, J.; Wang, X.; Schuurmans, D.; Bosma, M.; Xia, F.; Chi, E.; Le, Q.~V.; Zhou, D.; et~al. 2022.
\newblock Chain-of-thought prompting elicits reasoning in large language models.
\newblock \emph{Advances in neural information processing systems}, 35: 24824--24837.

\bibitem[{Wu et~al.(2024)Wu, Mu, Wu, Hou, Ma, Zhang, and Liu}]{wu2024voronav}
Wu, P.; Mu, Y.; Wu, B.; Hou, Y.; Ma, J.; Zhang, S.; and Liu, C. 2024.
\newblock Voronav: Voronoi-based zero-shot object navigation with large language model.
\newblock \emph{arXiv preprint arXiv:2401.02695}.

\bibitem[{Xia et~al.(2018)Xia, Zamir, He, Sax, Malik, and Savarese}]{xia2018gibson}
Xia, F.; Zamir, A.~R.; He, Z.; Sax, A.; Malik, J.; and Savarese, S. 2018.
\newblock Gibson env: Real-world perception for embodied agents.
\newblock In \emph{Proceedings of the IEEE conference on computer vision and pattern recognition}, 9068--9079.

\bibitem[{Yamagata, Khalil, and Santos-Rodriguez(2023)}]{yamagata2023q}
Yamagata, T.; Khalil, A.; and Santos-Rodriguez, R. 2023.
\newblock Q-learning decision transformer: Leveraging dynamic programming for conditional sequence modelling in offline rl.
\newblock In \emph{International Conference on Machine Learning}, 38989--39007. PMLR.

\bibitem[{Yang et~al.(2018)Yang, Wang, Farhadi, Gupta, and Mottaghi}]{yang2018visual}
Yang, W.; Wang, X.; Farhadi, A.; Gupta, A.; and Mottaghi, R. 2018.
\newblock Visual semantic navigation using scene priors.
\newblock \emph{arXiv preprint arXiv:1810.06543}.

\bibitem[{Ye et~al.(2021)Ye, Batra, Das, and Wijmans}]{ye2021auxiliary}
Ye, J.; Batra, D.; Das, A.; and Wijmans, E. 2021.
\newblock Auxiliary tasks and exploration enable objectgoal navigation.
\newblock In \emph{Proceedings of the IEEE/CVF international conference on computer vision}, 16117--16126.

\bibitem[{Yokoyama et~al.(2024)Yokoyama, Ha, Batra, Wang, and Bucher}]{yokoyama2024vlfm}
Yokoyama, N.; Ha, S.; Batra, D.; Wang, J.; and Bucher, B. 2024.
\newblock Vlfm: Vision-language frontier maps for zero-shot semantic navigation.
\newblock In \emph{2024 IEEE International Conference on Robotics and Automation (ICRA)}, 42--48. IEEE.

\bibitem[{Yu, Kasaei, and Cao(2023)}]{yu2023l3mvn}
Yu, B.; Kasaei, H.; and Cao, M. 2023.
\newblock L3mvn: Leveraging large language models for visual target navigation.
\newblock In \emph{2023 IEEE/RSJ International Conference on Intelligent Robots and Systems (IROS)}, 3554--3560. IEEE.

\bibitem[{Yue et~al.(2025)Yue, Chen, Lu, Zhao, Wang, Song, and Huang}]{yue2025does}
Yue, Y.; Chen, Z.; Lu, R.; Zhao, A.; Wang, Z.; Song, S.; and Huang, G. 2025.
\newblock Does reinforcement learning really incentivize reasoning capacity in llms beyond the base model?
\newblock \emph{arXiv preprint arXiv:2504.13837}.

\bibitem[{Zhai et~al.(2024)Zhai, Bai, Lin, Pan, Tong, Zhou, Suhr, Xie, LeCun, Ma et~al.}]{zhai2024fine}
Zhai, S.; Bai, H.; Lin, Z.; Pan, J.; Tong, P.; Zhou, Y.; Suhr, A.; Xie, S.; LeCun, Y.; Ma, Y.; et~al. 2024.
\newblock Fine-tuning large vision-language models as decision-making agents via reinforcement learning.
\newblock \emph{Advances in neural information processing systems}, 37: 110935--110971.

\bibitem[{Zhang et~al.(2025{\natexlab{a}})Zhang, Zhuang, Zhao, Ding, Lu, and Wang}]{zhang2025reinbot}
Zhang, H.; Zhuang, Z.; Zhao, H.; Ding, P.; Lu, H.; and Wang, D. 2025{\natexlab{a}}.
\newblock ReinboT: Amplifying Robot Visual-Language Manipulation with Reinforcement Learning.
\newblock \emph{arXiv preprint arXiv:2505.07395}.

\bibitem[{Zhang et~al.(2024{\natexlab{a}})Zhang, Wang, Wang, Li, Liu, Wei, Wang, Zhang, and Wang}]{zhang2024uni}
Zhang, J.; Wang, K.; Wang, S.; Li, M.; Liu, H.; Wei, S.; Wang, Z.; Zhang, Z.; and Wang, H. 2024{\natexlab{a}}.
\newblock Uni-navid: A video-based vision-language-action model for unifying embodied navigation tasks.
\newblock \emph{arXiv preprint arXiv:2412.06224}.

\bibitem[{Zhang et~al.(2024{\natexlab{b}})Zhang, Wang, Xu, Zhou, Hong, Fang, Wu, Zhang, and Wang}]{zhang2024navid}
Zhang, J.; Wang, K.; Xu, R.; Zhou, G.; Hong, Y.; Fang, X.; Wu, Q.; Zhang, Z.; and Wang, H. 2024{\natexlab{b}}.
\newblock Navid: Video-based vlm plans the next step for vision-and-language navigation.
\newblock \emph{arXiv preprint arXiv:2402.15852}.

\bibitem[{Zhang et~al.(2025{\natexlab{b}})Zhang, Hao, Xu, Zhang, Zhang, Wang, Zhang, Wang, Zhang, and Xu}]{zhang2025mapnavnovelmemoryrepresentation}
Zhang, L.; Hao, X.; Xu, Q.; Zhang, Q.; Zhang, X.; Wang, P.; Zhang, J.; Wang, Z.; Zhang, S.; and Xu, R. 2025{\natexlab{b}}.
\newblock MapNav: A Novel Memory Representation via Annotated Semantic Maps for Vision-and-Language Navigation.
\newblock \emph{arXiv preprint arXiv:2506.09839}.

\bibitem[{Zhang et~al.(2024{\natexlab{c}})Zhang, Zhang, Wang, Xiao, Jiang, Chen, and Xu}]{zhang2024trihelper}
Zhang, L.; Zhang, Q.; Wang, H.; Xiao, E.; Jiang, Z.; Chen, H.; and Xu, R. 2024{\natexlab{c}}.
\newblock TriHelper: Zero-Shot Object Navigation with Dynamic Assistance.
\newblock \emph{arXiv preprint arXiv:2403.15223}.

\bibitem[{Zheng et~al.(2024)Zheng, Huang, Zhao, Zhong, and Wang}]{zheng2024towards}
Zheng, D.; Huang, S.; Zhao, L.; Zhong, Y.; and Wang, L. 2024.
\newblock Towards learning a generalist model for embodied navigation.
\newblock In \emph{Proceedings of the IEEE/CVF Conference on Computer Vision and Pattern Recognition}, 13624--13634.

\bibitem[{Zhou, Hong, and Wu(2024)}]{zhou2024navgpt}
Zhou, G.; Hong, Y.; and Wu, Q. 2024.
\newblock Navgpt: Explicit reasoning in vision-and-language navigation with large language models.
\newblock In \emph{Proceedings of the AAAI Conference on Artificial Intelligence}, volume~38, 7641--7649.

\bibitem[{Zhou et~al.(2023)Zhou, Zheng, Pryor, Shen, Jin, Getoor, and Wang}]{zhou2023esc}
Zhou, K.; Zheng, K.; Pryor, C.; Shen, Y.; Jin, H.; Getoor, L.; and Wang, X.~E. 2023.
\newblock Esc: Exploration with soft commonsense constraints for zero-shot object navigation.
\newblock In \emph{International Conference on Machine Learning}, 42829--42842. PMLR.

\bibitem[{Zhuang et~al.(2024)Zhuang, Peng, Liu, Zhang, and Wang}]{zhuang2024reinformer}
Zhuang, Z.; Peng, D.; Liu, J.; Zhang, Z.; and Wang, D. 2024.
\newblock Reinformer: Max-return sequence modeling for offline rl.
\newblock \emph{arXiv preprint arXiv:2405.08740}.

\end{thebibliography}

\end{document}